# High-throughput relation extraction algorithm development associating knowledge articles and electronic health records

Yucong Lin[1,2], Keming Lu[3], Yulin Chen[4], Chuan Hong[5], Sheng Yu[1,2,6,*]

[1]*Center for Statistical Science, Tsinghua University, Beijing, China;*
[2]*Department of Industrial Engineering, Tsinghua University, Beijing, China;*
[3]*Department of Automation, Tsinghua University, Beijing, China;*
[4]*Department of Foreign Languages and Literatures, Tsinghua University, Beijing, China;*
[5]*Department of Biomedical Informatics, Harvard Medical School, Boston, MA, USA;*
[6]*Institute for Data Science, Tsinghua University, Beijing, China.*

*Correspondence to:
Sheng Yu
Weiqinglou Rm 209
Center for Statistical Science
Tsinghua University
Beijing, 100084, China
Email: syu@tsinghua.edu.cn
Tel: +86-10-62783842

**Abstract**

**Objective:** Medical relations are the core components of medical knowledge graphs that are needed for healthcare artificial intelligence. However, the requirement of expert annotation by conventional algorithm development processes creates a major bottleneck for mining new relations. In this paper, we present Hi-RES, a framework for high-throughput relation extraction algorithm development. We also show that combining knowledge articles with electronic health records (EHRs) significantly increases the classification accuracy.

**Methods:** We use relation triplets obtained from structured databases and semistructured webpages to label sentences from target corpora as positive training samples. Two methods are also provided for creating improved negative samples by combining positive samples with naïve negative samples. We propose a common model that summarizes sentence information using large-scale pretrained language models and multi-instance attention, which then joins with the concept embeddings trained from the EHRs for relation prediction.

**Results:** We apply the Hi-RES framework to develop classification algorithms for disorder-disorder relations and disorder-location relations. Millions of sentences are created as training data. Using pretrained language models and EHR-based embeddings individually provides considerable accuracy increases over those of previous models. Joining them together further tremendously increases the accuracy to 0.947 and 0.998 for the two sets of relations, respectively, which are 10-17 percentage points higher than those of previous models.

**Conclusion:** Hi-RES is an efficient framework for achieving high-throughput and accurate relation extraction algorithm development.

# 1 Introduction

The relations between medical concepts are essential for medical ontologies and knowledge graphs, and with the rapid development of artificial intelligence in healthcare, which increasingly relies on these high-level infrastructures [1–3], the methodology for identifying medical relations must also improve.

Until today, pattern-based methods have been popularly used to extract relations from medical texts. For example, the hierarchical semantic relation [X, is a kind of, Y] is a simple relation that can be effectively extracted by patterns, which are generally implemented as finite state machines [4] and regular expressions in particular. The simplicity of these patterns makes them easy to create and implement at scale [5,6]. However, the simplicity of pattern-based methods also limits their accuracy [5] and makes them ineffective for capturing relations expressed in sophisticated sentences. Machine learning can create models that can adapt to diverse expressions if they are given enough annotated training data. The models can use various features for learning, including bag-of-words (BOW), parts-of-speech (POS), the relative positions of words, semantic relations from WordNet [7], parse trees [8–11], and embeddings, which is an important development in recent years [12–14]. Support vector machines with dependency tree-based kernels achieved the best accuracy among conventional machine learning methods [9–11,15]. Deep learning introduced entirely new techniques and revolutionized the field. Liu *et al*. introduced convolutional neural networks (CNNs) for relation extraction [16]. Several studies found that recurrent neural networks (RNNs), which were designed to process sequential inputs, such as languages, achieved better results than those of CNNs [17,18]. The attention mechanism [19] that demonstrated success in machine translation has also been incorporated into relation extraction and has attained improved performance without using traditional natural language processing (NLP) features, such as POS and WordNet, further demonstrating the potential of automatic feature extraction with deep learning [20]. Various other novel architectures have also been proposed [21–26]. Until now, relation extraction models based on the Transformer architecture [27], especially the pretrained BERT model [28], have achieved state-of-the-art performance levels [29–31].

Recalling the goal of using algorithms to automatically extract various kinds of relations to support advanced artificial intelligence in healthcare, one easily realizes that the main bottleneck has shifted from the capability of the models to the lack of annotated training data. Annotated medical data are famously scarce due to both privacy protection requirements and the tremendous demand for experts' time. Driving forces in the field, such as i2b2, n2c2, and OHNLP, undertake shared tasks to provide annotated data for NLP research [32,33], but the sample sizes are small, especially for deep learning, and the annotations are for only a few relation types and text forms. As a result, researchers repeatedly improved relation extraction models using the same relations and texts but could not apply the models to mine new relations from new texts. To truly achieve the goal of relation mining, a new mode of high-throughput relation extraction is needed.

High-throughput technology has been achieved in multiple biomedical fields and has revolutionized them. For example, high-throughput drug screening has allowed researchers to perform quick tests among millions of chemical compounds [34,35]; high-throughput phenotyping has enabled unsupervised extraction of thousands of phenotypes simultaneously from electronic health records (EHRs) with comparable accuracy to those of supervised algorithms [36–38]. A high-throughput mode for relation extraction should allow us to develop an extraction algorithm relatively easily for a new relation, a process that would typically require that no manually labeled training samples are used. A training sample for relation extraction includes three elements: a text piece stating the relation between two entities, the marked positions of the two entities, and a label for the classification of the relation. The distant supervision mechanism (also known as weak supervision in many fields) can automatically label data, making high-throughput relation extraction possible [39]. Under the assumption of distant supervision, if a sentence contains two entities with a known relation (e.g., from an existing knowledge base), then the sentence is labeled as expressing that relation. For example, if we know the relation [diabetes, may cause, weight loss], we can create a training sample from any sentence that contains both "diabetes" and "weight loss". Several deep neural networks have been proposed using distant supervision [40,41]. Lin *et al*. extended the extraction range and improved the accuracy of disorder-disorder relations [42]. The workflow of Lin *et al*. exhibits much potential for generalizability and standardization. Based on this, we propose the High-throughput Relation Extraction System (Hi-RES) framework for algorithm development, which involves the collection of relation triplets, generation of positive and negative samples, and formation of a common model. We follow the Hi-RES framework to develop algorithms for disorder-disorder and disorder-anatomy relations for demonstration purposes.

In this paper, we also show that associating knowledge articles (such as Wikipedia articles, research papers, and textbooks) with EHRs can effectively improve algorithm accuracy. Knowledge articles and EHRs present information from different dimensions: the former directly states entity relations through language, while the latter is good at revealing associations via cooccurrence that can be computed at various windows. It has been shown that combining the two sources can benefit informatics tasks. For example, Zhao and Weng joined EHR data and PubMed abstracts to extract weighted risk factors for the prediction of pancreatic cancer [43], and Liu *et al*. achieved human-level performance in abbreviation expansion in clinical texts by learning word embeddings from PubMed, PMC, and biomedical Wikipedia articles [44]. EHR data have also been used as a sole source of data for relation extraction, such as mining side effects of drugs [45–50], phenotype-genotype associations [51–53], and clinical temporal relation extraction [25,26,54,55]. However, to the extent of our knowledge, there are few studies on jointly mining knowledge articles and EHRs for relation extraction. We present a simple way to merge the free text information from knowledge articles and the cooccurrence information from EHRs into the Hi-RES framework with tremendous benefit to the accuracy of the algorithm.

## 2 Methods

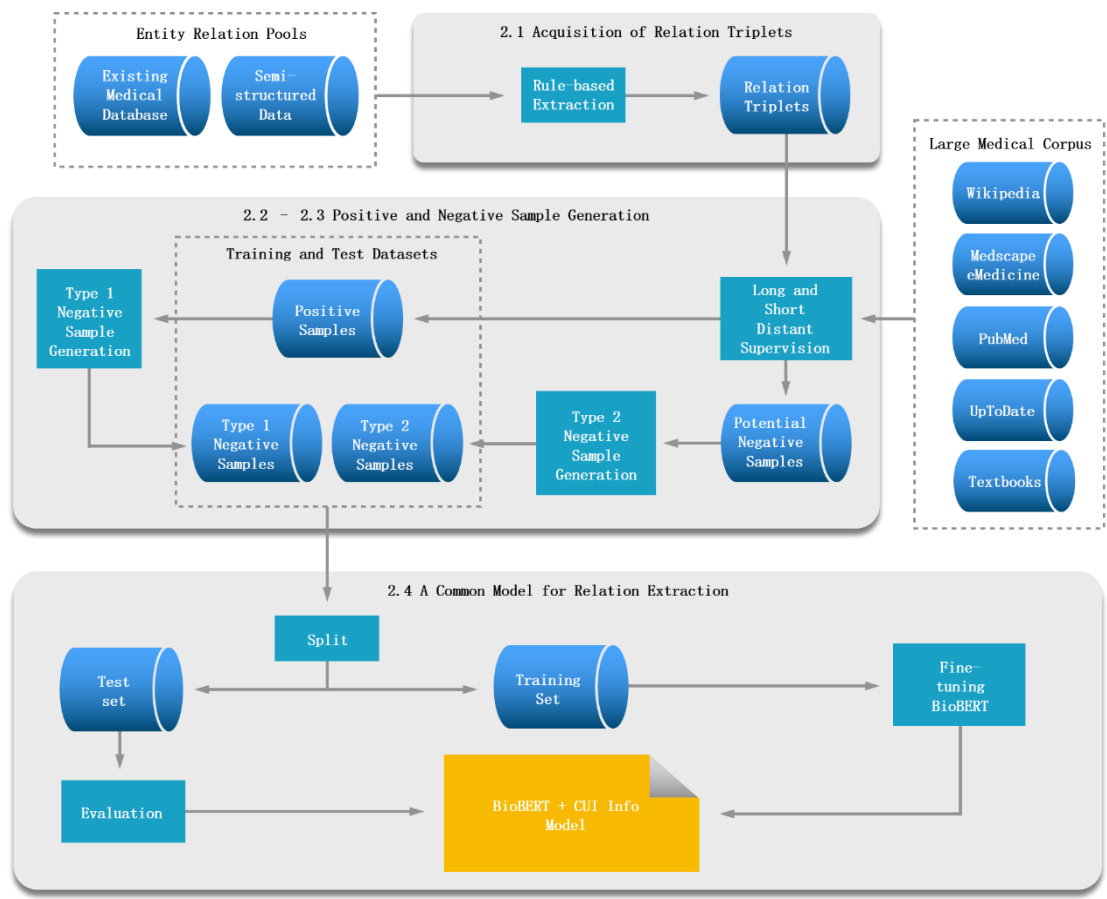

Figure 1: The pipeline for data processing and model training.

In this section, we introduce the sample generation workflow and the model architecture of Hi-RES. Figure 1 illustrates the interactions among the various components of Hi-RES.

**2.1 Acquisition of relation triplets**

Acquiring relation triplets is the first step in the preparation of training data. A relation triplet is a representation of a relation that consists of a head entity, a tail entity, and the relation between them. For example, a triplet can be [diabetes, may cause, weight loss], where in practice, the head and tail entities should be normalized to ontology concepts, e.g., by using Concept Unique Identifiers (CUIs) from the Unified Medical Language System (UMLS) [56] instead of free text. Relation triplets are the final products that we want to obtain from relation extraction, but in distant supervision, we also need initial triplets to generate positive training samples.

Some ideal sources for obtaining relation triplets are the existing structured knowledge bases, such as the UMLS [56], DBpedia [57], and Wikidata [58]. The benefit of using these knowledge bases is that the entities and relations are already normalized and structured, making them easy to use. However, the relations of interest may not be available from existing knowledge bases. In this case, one can alternatively extract relation triplets from semistructured content on the Internet. Basic relations have been collected by many websites. The pages of the website usually follow a very standard template, which allows one to easily locate sections about the target relations, where the entities are usually presented in lists and tables. Figure 2 shows an example of such a page, which lists lab tests for rheumatoid arthritis. Most commonly, the page title provides the head entity, the section title specifies the relation, and the lists/tables in the section provide the tail entities. Therefore, we can write simple web scraping scripts and apply named entity recognition (NER) to extract the entities, identify them with UMLS CUIs, and assemble them into relation triplets. The reason why we target lists and tables is that they have simple semantics, where each entry is usually an entity, which leads to greatly reduced error rates when interpreting the meaning of entries and creating triplets. However, there are times when an entry can be complex (typically signaled by its length), such as the last entry of the list in Figure 2. One may want to avoid applying NER on such entries to avoid potential errors. To create sufficient training samples for deep learning, we suggest extracting at least thousands of triplets in this step.

Figure 2: A webpage presenting lab tests for rheumatoid arthritis in the form of a list (https://emedicine.medscape.com/article/331715-workup, accessed on July 28, 2020).

**2.2 Positive sample generation**

The collected relation triplets allow us to create positive samples using the distant supervision mechanism. Here, by positive samples, we mean samples that express any target relation, which can include multiple classes if we aim to extract multiple relations at once. Conversely, negative samples refer to samples that do not express the target relations.

Training samples are generated using sentences from the target corpus, which does not have to be the websites where the relation triplets were collected. Any sentence that contains both entities of a triplet is tagged with the relation of that triplet. Similar to Lin *et al*. [42], we also consider long-distance expressions where one entity is the article title and the other is part of a sentence. For both long and short distances, the section headings that lead to the sentence are extracted and considered as part of the sample, as they usually carry important contextual information about the relation that is not repeated in the sentence [42]. Finally, we use sentence-level attention to mitigate the errors from distant supervision labeling [41]. Therefore, the labeled sentences generated from the same triplet are grouped and considered as one sample. This will be further explained in Section 2.4.

**2.3 Negative sample generation**
Negative samples are needed to show the classifier what kinds of sentences do not express target relations, and their quality can significantly affect the classification accuracy. Lin *et al*. generated negative samples by choosing sentences that contained entities whose semantic types were irrelevant to the target relations [42]. These samples (referred to as "potential negative samples" hereafter) are guaranteed to be negative samples due to their incompatible entity types. However, they are also too different from the positive samples and too easy to distinguish, making the classifier insufficiently discriminative when applied to predict new relations. Ideal negative samples should be as similar to the positive ones as possible so that the classifier is forced to learn difficult situations to determine the right classification boundaries. Here, we propose two types of negative samples based on the original sample.

The first type of negative samples, called "Type 1 negative samples", are obtained by replacing the words between the two entities of a positive sentence. A sentence labeled positive can be divided into three parts by the two entities: [Head Text] [E1] [Middle Text] [E2] [Tail Text], where each part is allowed to be empty. We randomly select a potential negative sample sentence, which can be similarly divided into three parts. To create a negative sentence, we replace the [Middle Text] component of the positive sentence with that of the potential negative sentence, and we also replace [E1] and [E2] with random entities of their corresponding semantic types. Therefore, a negative sample possesses the following structure: [Head Text] [E1-Replaced] [Middle Text-Replaced] [E2-Replaced] [Tail Text]. Since the middle text is usually the most informative part for expressing relations in a sentence, replacing it with the corresponding part of the potential negative sentence not only ensures that the target relation is not expressed but also maintains a certain level of similarity to the positive sentences. The reason for replacing the entities will be explained in Section 2.4.

The "Type 2 negative samples" are selected from the potential negative samples by evaluating their similarity to the positive samples. Potential negative samples are grouped by the involved entity pairs in the same way as the positive samples. We use word embedding to evaluate similarity. Skip-gram-based word embeddings [13] are trained on more than 28 GB of free-text medical corpora, including Wikipedia, UpToDate, Medscape eMedicine, PubMed abstracts, and medical textbooks. The embedding vector of a sentence, denoted by $S^{sg}$, is taken as the simple average of the sentence's words-embedding vectors, where the superscript $sg$ indicates that the representation is derived from the Skip-gram embeddings. Weighted averages, such as the inverse document frequency method, are not used because common words can be important for expressing relations and should not receive reduced weights. Furthermore, for a sample that contains multiple sentences (recall that a sample comprises sentences grouped by the same involved entity pair), the sample embedding is the average of the sentence embeddings, denoted by $E^{sg}$. Thus, we can define the similarity between two samples or entity pairs through the cosine similarity:

$$Similarity\left(E_i^{sg}, E_j^{sg}\right) = \frac{\langle E_i^{sg}, E_j^{sg} \rangle}{||E_i^{sg}|| \cdot ||E_j^{sg}||}.$$

To create Type 2 negative samples, we first obtain a large number of potential negative samples. Each sample is evaluated with all the positive samples for similarity, and the highest similarity value is kept. We then sort the potential negative samples by similarity in descending order and select the most similar ones as negative samples, selecting the same amount as the number of positive samples.

**2.4 A common model for relation extraction**

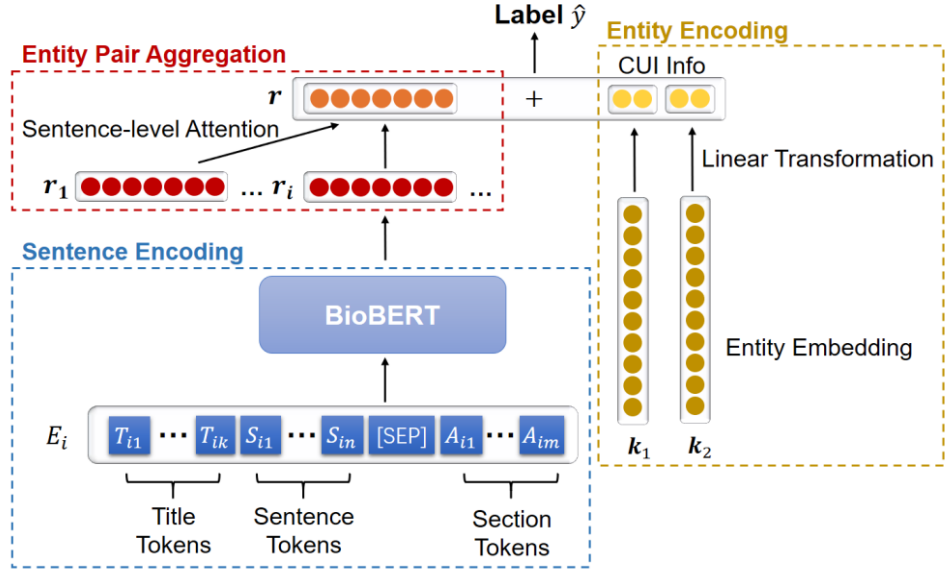

Figure 3. The model structure

We use a modularized design for the relation extraction model so that components can be easily replaced when the corresponding technologies improve. As Figure 3 shows, the model is composed of three parts: a sentence encoder that converts input sentences to vectors, an aggregator that combines sentence vectors of the same pair of entities into a single vector, and an entity encoder that encodes the involved entities' corresponding CUIs from EHR-based embeddings.

The first part is the sentence encoder. We extend the work of Lin *et al*. [42] and include the article title and section headings as part of the sentence for long-distance extraction, and we replace the earlier gated recurrent unit (GRU) modules with the more advanced BioBERT [59]. For the *i*-th sentence of a sample, we use the BioBERT tokenizer to tokenize the title as the sequence $T_i$, the sentence as $S_i$, and the section headings as $A_i$ (multilevel headings are first concatenated), and the sequences are concatenated as $E_i$:

$$E_i = \text{concat}(T_i, S_i, [\text{SEP}], A_i),$$

where [SEP] is BioBERT's reserved separator token. The entity names are masked by tokens representing the corresponding semantic types, which can force the model to learn from the sentence patterns instead of by memorizing entity names. The sequence $E_i$ is then fed to BioBERT for encoding:

$$\boldsymbol{r}_i = \text{BioBERT}(E_i),$$

where $\boldsymbol{r}_i \in \mathbb{R}^{d_r}$ is obtained by using attention to aggregate the hidden vectors from the last layer of BioBERT. Ideally, BioBERT should be trained as part of the model. However, due to limited computing resources, we fine-tune BioBERT before training the model by directly using the individual sentences to predict their labels.

The second part of the model is a sentence vector aggregator. Since one entity pair may contain numerous sentences, to reduce the computational cost, the model randomly selects $n_s$ sentences from each entity pair and combines their embeddings into one using the attention mechanism [41]:

$$R = [\boldsymbol{r}_1, \boldsymbol{r}_2, \ldots, \boldsymbol{r}_{n_s}],$$
$$\boldsymbol{\alpha}_{EP} = \text{softmax}(R^T \boldsymbol{v}_{EP}),$$
$$\boldsymbol{r} = R\boldsymbol{\alpha}_{EP},$$

where $\boldsymbol{v}_{EP} \in \mathbb{R}^{d_r}$ is a trained query vector, and $\boldsymbol{\alpha}_{EP}$ is the computed weight vector for the sentences. The attention mechanism can automatically increase the weights of sentences that express the labeled relations and decrease the weights of those that do not, making the model more robust to incorrectly labeled sentences brought by distant supervision. $\boldsymbol{r} \in \mathbb{R}^{d_r}$ is the final vector representation of the entity pair obtained by aggregating over the representations of sentences.

The third part of the model is the entity encoder. The entities are masked in the sentences to prevent overfitting. Here, entity information is brought back into the model as abstract embeddings obtained from EHR data, which provide important association information about the two entities. It is important to note that in generating Type 1 negative samples, we have replaced the entities from the positive samples with random entities of the same semantic type. The reason is that if we kept the original entities in the generated negative samples, the entity encoder would be given the impression that entity embeddings were uninformative because related entities appeared in both positive and negative samples. Therefore, it is necessary to use random entities in the negative samples to allow the entity encoder to infer from the embeddings.

Since we have normalized the entities to UMLS CUIs, we use cui2vec [60] to provide EHR-based CUI embeddings. When a CUI is not covered by cui2vec, we first attempt to find its hypernym CUI's embedding along the UMLS hierarchy, and if the hypernym is not covered, we use the average of all the known CUI embeddings as the CUI's embedding. The embeddings (referred to as the CUI Info in later model comparisons) are projected to lower dimensions and concatenated with the sentence embedding for the final classification.

$$\boldsymbol{c}_j = W_j \boldsymbol{k}_j + \boldsymbol{b}_j, \text{ for } j = 1,2,$$

$$f = \text{concat}(r, c_1, c_2),$$
$$\hat{y} = \text{softmax}(Wf + b),$$

where $k_j \in R^{d_k}$ stands for the CUI embedding of the $j$-th entity, $c_j \in R^{d_c}, j = 1,2$ are the embeddings' low-dimensional projections, and $\hat{y}$ is the predicted distribution used for classification.

## 3 Experiments

To demonstrate the use and effectiveness of Hi-RES, we developed extraction algorithms for two sets of relations. The first set of relations were among entities of the UMLS Disorder (DISO) semantic group, including the undirected relation "differential diagnosis" (DDx) and the directed relations "may cause" (MC) and "may be caused by" (MBCB). These relations were the same as those of Lin *et al*. [42], but the samples were different and largely expanded. The other set involved the directed relation "occurs in" (IN) between a DISO entity and a UMLS Anatomy (ANAT) entity.

The initial relation triplets were collected from multiple sources. For the DISO-ANAT relations, we collected the IN triplets from the UMLS Metathesaurus relation table MRREL. We also extended the IN relations using the following rule: if [D, occurs in, $A_1$] and [$A_1$, part_of / anatomic_structure_is_physical_part_of / anatomic_structure_has_location, $A_2$], then [D, occurs in, $A_2$], where the hierarchical relations of ANAT-ANAT were also found in MRREL. The extension increased the number of IN triplets by 148%, from 40,729 to 100,808. For the DISO-DISO relations, we extracted from the structured Diseases Database [61] and the semistructured webpages of Medscape eMedicine [62] and obtained 11244 DDx triplets and 14021 MC/MBCB triplets. Additional customized data cleaning processes were applied, such as deleting triplets if either of the entities belonged to the "Finding" semantic type because this subclass contained too many irrelevant concepts and terms, such as Male (C0024554) and Marriage (C0024841), which would cause a large number of false discoveries.

The medical corpora from which we created positive and negative samples included Wikipedia, Medscape eMedicine, paper abstracts from PubMed, and four textbooks - *Harrison's Principles of Internal Medicine 20th Edition, Kelley's Textbook of Internal Medicine 4th Edition, Sabiston Textbook of Surgery: The Biological Basis of Modern Surgical Practice 20th Edition, and Kumar and Clark's Clinical Medicine 7th Edition* [63–66]. The total size of these corpora exceeded 28 GB, comprising over 20 million medicine-related articles. We created positive samples using the method described in Section 2.2. Among the corpora, PubMed was treated differently because its abstracts did not contain informative section headings, and its titles were usually sentences rather than medical entities. Therefore, we regarded PubMed's article titles as sentences and set the title and section headings of samples as empty to conform with our data structure. Table 1 summarizes the number of collected positive samples. The Supplementary Material provides statistics for sentences from each source corpus.

Table 1: The number of collected relation triplets, positive samples (i.e., triplets whose entities were matched with sentences), and involved sentences.

|  | DISO-DISO | | | | DISO-ANAT |
|---|---|---|---|---|---|
|  | **Total** | **DDx** | **MC** | **MBCB** | **IN** |
| Collected relation triplets | 36,157 | 12,599 | 14,243 | 9,315 | 100,808 |
| Positive samples | 19,748 | 9,372 | 5,558 | 4,818 | 5,901 |
| Sentences in positive samples | 1,128,400 | 356,079 | 216,399 | 185,859 | 155,313 |

With the positive samples collected, we generated Type 1 and Type 2 negative samples following the two approaches listed in Section 2.3. Three datasets were created to evaluate the effectiveness of the artificial negative samples. Dataset-1 and Dataset-2 contained Type 1 and Type 2 negative samples, respectively, and Dataset-mix contained half of the Type 1 negative samples from Dataset-1 and half of the Type 2 negative samples from Dataset-2. The three datasets shared the same positive samples, and the ratio of positive and negative samples was approximately 1:1 in each. The datasets were randomly split 80:20 for training and testing. The sets were separated by entity pair to prevent information leaks, i.e., sentences about the same pair of entities either all go to the training set or all go to the test set.

To evaluate the performance of the model proposed in Section 2.4, we employed a series of baseline models for comparison, including (a) BOW Naïve Bayes, (b) BOW support vector machines (SVM), (c) CNN, (d) Bi-GRU, (e) Bi-GRU + attention + article structures [42], (f) Bi-GRU + attention + article structures + CUI Info, (g) CUI only, (h) BERT, (i) BERT + CUI Info, and (j) BioBERT + CUI Info. Models (g) and (h) corresponded to the entity encoder and the sentence encoder introduced in Section 2.4, respectively, and they served as ablation tests. Note that (h) differed from (c) and (d) in that we included article structure information in (h) but not in (c) and (d). Both (i) and (j) were the same proposed model but (i) used BERT instead of BioBERT as the sentence encoder. The models were also different in terms of how the multiple sentences of an entity pair were used for training and prediction. Models (a)-(d) and (h) used the original distant supervision mechanism for training and majority voting for prediction. Models (e), (f), (i), and (j) incorporated attention to weight the samples by credibility during training and prediction. The hyperparameters for the proposed models (i) and (j) mainly followed the settings in used previous papers [17,29,67] (Supplementary Material).

## 4 Results

Table 2: Accuracy, recall, precision, and F-score values for DISO-ANAT relations.

| Model | Dataset-1 | | | | Dataset-2 | | | | Dataset-mix | | | |
|---|---|---|---|---|---|---|---|---|---|---|---|---|
|  | Accuracy | Recall | Precision | F-score | Accuracy | Recall | Precision | F-score | Accuracy | Recall | Precision | F-score |
| a. Naïve Bayes | 0.677 | 0.724 | 0.651 | 0.686 | 0.828 | 0.688 | 0.976 | 0.807 | 0.759 | 0.693 | 0.820 | 0.751 |
| b. SVM | 0.651 | 0.616 | 0.650 | 0.633 | 0.836 | 0.728 | 0.951 | 0.825 | 0.750 | 0.685 | 0.808 | 0.741 |
| c. CNN | 0.771 | 0.801 | 0.747 | 0.773 | 0.918 | 0.897 | 0.943 | 0.919 | 0.825 | 0.846 | 0.825 | 0.835 |
| d. Bi-GRU | 0.867 | 0.862 | 0.865 | 0.863 | 0.931 | 0.923 | 0.944 | 0.933 | 0.868 | 0.861 | 0.884 | 0.872 |
| e. + attention + article structures | 0.886 | 0.897 | 0.873 | 0.885 | 0.934 | 0.914 | 0.958 | 0.935 | 0.887 | 0.897 | 0.888 | 0.892 |
| f. + CUI Info | 0.936 | 0.955 | 0.917 | 0.936 | 0.968 | 0.958 | 0.980 | 0.969 | 0.940 | 0.935 | 0.950 | 0.942 |
| g. CUI only | 0.990 | 0.991 | 0.908 | 0.948 | 0.998 | 0.992 | 0.995 | 0.993 | 0.996 | 0.990 | 0.993 | 0.991 |
| h. BERT | 0.965 | 0.971 | 0.959 | 0.965 | 0.981 | 0.985 | 0.978 | 0.981 | 0.990 | 0.990 | 0.990 | 0.990 |
| i. BERT + CUI Info | 0.994 | 0.997 | 0.990 | 0.993 | 0.997 | **0.998** | 0.996 | 0.997 | 0.993 | 0.997 | 0.990 | 0.993 |
| j. BioBERT + CUI Info | **0.998** | **0.998** | **0.998** | **0.998** | **0.998** | 0.997 | **0.998** | 0.997 | **0.997** | **0.998** | **0.997** | **0.997** |

Table 3: Overall and positive sample accuracy values for DISO-DISO relations.

| Model | Dataset-1 | | Dataset-2 | | Dataset-mix | |
|---|---|---|---|---|---|---|
| | Overall Acc. | Positive Acc. | Overall Acc. | Positive Acc. | Overall Acc. | Positive Acc. |
| a. Naïve Bayes | 0.509 | 0.075 | 0.593 | 0.289 | 0.504 | 0.154 |
| b. SVM | 0.406 | 0.330 | 0.566 | 0.322 | 0.483 | 0.317 |
| c. CNN | 0.755 | 0.549 | 0.761 | 0.585 | 0.737 | 0.557 |
| d. Bi-GRU | 0.828 | 0.678 | 0.821 | 0.690 | 0.822 | 0.698 |
| e. + attention + article structures | 0.826 | 0.725 | 0.839 | 0.731 | 0.819 | 0.728 |
| f. + CUI Info | 0.896 | 0.830 | 0.900 | 0.833 | 0.894 | 0.835 |
| g. CUI only | 0.785 | 0.691 | 0.782 | 0.724 | 0.780 | 0.703 |
| h. BERT | 0.895 | 0.803 | 0.905 | 0.849 | 0.904 | 0.840 |
| i. BERT + CUI Info | 0.934 | 0.876 | 0.933 | 0.888 | 0.928 | 0.881 |
| j. BioBERT + CUI Info | **0.947** | **0.899** | **0.941** | **0.911** | **0.944** | **0.909** |

Table 2 and Table 3 show the performance of the models on classifying DISO-ANAT and DISO-DISO relations. Recall and precision values for individual DISO-DISO relations are given in the Supplementary Material. The deep learning algorithms are substantially superior to the traditional machine learning methods for both tasks. Model (e), Bi-GRU + attention + article structures, is the main baseline from the previous study [42]. Its performance increase over that of simple Bi-GRU was not as significant as in [42] due to the dominating proportion of PubMed abstracts in the new datasets, which did not have article structures. Adding CUI Info in (f) brought tremendous improvement over (e). Indeed, CUI alone was already powerful for DISO-ANAT relations but not for DISO-DISO relations. Only using the sentence encoder (h) also resulted in good performances on DISO-ANAT relations. For DISO-DISO relations, although (h) was not sufficiently accurate, its performance was nevertheless significantly better than that of the baseline (e). Drastic improvements occurred when we joined the knowledge article information with the information from the EHRs. The advantage of (i) over (g) and (h) proves that the two sources of information are not repetitive but rather complement each other. Finally, the advantage of (j) over (i) shows that pretraining based on a biomedical corpus does improve performance in medical informatics tasks. Overall, the proposed model (j) achieved near-perfect accuracy on DISO-ANAT relations and a 10-17 percentage point accuracy improvement on DISO-DISO relations over that of (e), bringing the overall accuracy to 94-95% and the positive sample accuracy to 90-91%. Improvement of such magnitude is very rare in NLP tasks.

Table 4: Cross-testing of the models on alternate negative samples. Columns indicate training sets, and rows indicate test sets.

| Testing Samples | DISO-ANAT Model | | | DISO-DISO Model | | |
|---|---|---|---|---|---|---|
| | Dataset-1 | Dataset-2 | Dataset-mix | Dataset-1 | Dataset-2 | Dataset-mix |
| Neg. 1 | --- | 0.96 | 0.997 | --- | 0.862 | 0.904 |
| Neg. 2 | 0.987 | --- | 0.999 | 0.974 | --- | 0.917 |

To evaluate the "authenticity" of the simulated negative samples, we cross-tested the fitted proposed models on alternate test data because good generalizability to a different type of sample would indicate high quality sample simulation. Table 4 shows the results of testing the model trained on Dataset-1 on the test data of Type 2 negative samples and vice versa and testing the model trained on Dataset-mix on both negative sample sets. The table shows that for both sets of relations, the models trained on Dataset-1 could generalize well to Type 2 negative samples, but the reverse direction was weaker. Mixing the two types of negative samples during training appears to be a good hedging strategy.

## 5 Discussion

A high-throughput relation extraction framework, such as Hi-RES, is the result of various related technologies, including distant supervision, attention mechanisms, large-scale pretrained language models, and representation learning. The advent of these technologies and their combination truly made the high-throughput mode possible for relation extraction, as it did not take much effort for us to expand to the new DISO-ANAT relation from existing pipelines [42] with nearly 100% accuracy. The vast amount of online content is another key element that enables Hi-RES by providing both seed relation triplets and sentences for training. As a result, by aggregating online information sources ranging from knowledge articles to research papers, we were able to acquire over 1 million sentences to serve as training data, which is not only essential for today's deep learning models but is also critical to counter the noise found in weakly labeled training data.

In Hi-RES, we also provided improved solutions to generate negative samples, which are equally important as positive samples. Good negative samples should resemble sentences that contain two entities of proper semantic types but do not express target relations. It is important to note that even using reserved test data, as in Table 2 and Table 3, is insufficient for evaluating the authenticity/usefulness of the simulated negative samples, as they follow the same pattern as those in the training data. In fact, naïve negative samples result in increased accuracy, while good negative samples should make the training appropriately difficult. Cross-testing for the generalizability of samples generated by alternative methods is one way to evaluate negative samples meaningfully. As Table 4 shows, Type 1 negative samples appeared to be better than Type 2 negative samples in terms of generalization. Mixing the two types of samples for training is also a reasonable choice.

To provide an easy-to-use common model, Hi-RES did not adopt the graph information from the previous work in [42], not only because the previous graph features could not generalize to other relations but also because graphs contain the inherent "cold-start" problem that causes the prediction capability to drop on new entities that possess few existing links in the graph. Instead, we incorporated EHR-based concept embeddings, which dramatically improved the prediction accuracy. The extent of the improvement on the old DISO-DISO relations was far greater than that which graph information could achieve in [42] (Model (f) over (e) in Table 3). More importantly, the benefit of the EHR-based concept embeddings stacks on top of the already powerful pretrained language models ((i), (j) over (h)), and together, they have raised the accuracy of relation extraction to a whole new level.

We conducted a manual review of the prediction quality. Although the accuracy of DISO-ANAT relations was nearly 100%, some predicted relations were overly general and useless. For instance, the model predicted [giant cell tumor of bone, occurs in, body], which is correct but useless. This was caused by overextending the basic UMLS-recorded relations described in Section 3 and can be prevented by rule-based cleaning. The DISO-DISO relations were more difficult to predict than the DISO-ANAT relations. We found that part of the difficulty could be attributed to differing understandings from professionals. We randomly selected 100 DISO-DISO entity pairs from the test set and two M.D. candidates manually annotated their relations. The Cohen's kappa coefficient of the two annotators was 0.757, showing only medium consistency, which indicates that information from online sentences could be inconsistent as well. Cue words in some sentences also misled some predictions. For example, "*The differential diagnosis includes unusual infections such as relapsing fever, malignancy and premalignant states (Schnitzler syndrome), cyclic neutropenia, and systemic juvenile idiopathic arthritis (SJIA) /adult-onset still's disease (AOSD).*", a sentence from UpToDate, caused all models to predict [relapsing fevers, DDx, Schnitzler syndrome], though the truth was MBCB. Hi-RES possesses a modular design, and we hope that new progress in the pretraining of language models, such as GPT-3 [68], can help distinguish these difficult sentences.

EHR-based embedding is another modular part of Hi-RES that we expect to improve in the future. In this paper, we only experimented in a preliminary way (though very successfully) by linearly transforming and concatenating the embeddings for later prediction. Other more complex transformations may prove more effective. It is also important to note that there are various ways to generate embeddings from EHRs. In addition to changing models, one can also adjust the cooccurrence window and alter which data are used. For example, the prediction of disease relations with laboratory tests and procedures may benefit from an embedding method that focuses more on structured data than on free text. How to more effectively use EHR data is an important topic for investigation.

## 6 Conclusion

In this paper, we proposed Hi-RES, a framework for high-throughput medical relation extraction algorithm development without requiring expert annotation, which makes it possible to quickly create highly accurate machine learning algorithms to mine new relations that facilitate the construction of knowledge graphs. The combination of EHR-based embeddings with knowledge articles, the utilization of large-scale pretrained language models, and the use of large datasets made available by the high-throughput mode tremendously raised the accuracy of the model from previous levels. The modular design of Hi-RES allows for the performance of agile experiments for improvement. In particular, more in-depth ways of creating and using EHR-based embeddings warrant further research.

## References

1 Goodwin T, Harabagiu SM. Automatic generation of a qualified medical knowledge graph and its usage for retrieving patient cohorts from electronic medical records. In: *Semantic Computing (ICSC), 2013 IEEE Seventh International Conference on*. IEEE 2013. 363–370.

2 Li X, Wang Y, Wang D, *et al.* Improving rare disease classification using imperfect knowledge graph. *BMC Medical Informatics and Decision Making* 2019;**19**:238.

3 Hasan SS, Rivera D, Wu X-C, *et al.* Knowledge graph-enabled cancer data analytics. *IEEE journal of biomedical and health informatics* 2020;**24**:1952–1967.

4 Arnold P, Rahm E. Extracting semantic concept relations from wikipedia. In: *Proceedings of the 4th International Conference on Web Intelligence, Mining and Semantics (WIMS14)*. 2014. 1–11.

5 Kilicoglu H, Fiszman M, Rodriguez A, *et al.* Semantic MEDLINE: a web application for managing the results of PubMed Searches. In: *Proceedings of the third international symposium for semantic mining in biomedicine*. Citeseer 2008. 69–76.

6 Rindflesch TC, Kilicoglu H, Fiszman M, *et al.* Semantic MEDLINE: An advanced information management application for biomedicine. *Information Services & Use* 2011;**31**:15–21.

7 Miller GA. WordNet: a lexical database for English. *Communications of the ACM* 1995;**38**:39–41.

8 Miwa M, Bansal M. End-to-End Relation Extraction using LSTMs on Sequences and Tree Structures. In: *Proceedings of the 54th Annual Meeting of the Association for Computational Linguistics (Volume 1: Long Papers)*. 2016. 1105–1116.

9 Zelenko D, Aone C, Richardella A. Kernel methods for relation extraction. *Journal of machine learning research* 2003;**3**:1083–1106.

10 Bunescu RC, Mooney RJ. A shortest path dependency kernel for relation extraction. In: *Proceedings of the conference on human language technology and empirical methods in natural language processing*. Association for Computational Linguistics 2005. 724–731.

11 Culotta A, Sorensen J. Dependency tree kernels for relation extraction. In: *Proceedings of the 42nd annual meeting on association for computational linguistics*. Association for Computational Linguistics 2004. 423.

12 Turian J, Ratinov L, Bengio Y. Word representations: a simple and general method for semi-supervised learning. In: *Proceedings of the 48th annual meeting of the association for computational linguistics*. Association for Computational Linguistics 2010. 384–394.

13 Mikolov T, Sutskever I, Chen K, *et al.* Distributed representations of words and phrases and their compositionality. In: *Advances in neural information processing systems*. 2013. 3111–3119.

14 Pennington J, Socher R, Manning C. Glove: Global vectors for word representation. In: *Proceedings of the 2014 conference on empirical methods in natural language processing (EMNLP)*. 2014. 1532–1543.

15 Hoffmann R, Zhang C, Ling X, *et al.* Knowledge-based weak supervision for information

extraction of overlapping relations. In: *Proceedings of the 49th Annual Meeting of the Association for Computational Linguistics: Human Language Technologies-Volume 1*. Association for Computational Linguistics 2011. 541–550.

16  Liu C, Sun W, Chao W, *et al.* Convolution neural network for relation extraction. In: *International Conference on Advanced Data Mining and Applications*. Springer 2013. 231–242.

17  Cai R, Zhang X, Wang H. Bidirectional recurrent convolutional neural network for relation classification. In: *Proceedings of the 54th Annual Meeting of the Association for Computational Linguistics (Volume 1: Long Papers)*. 2016. 756–765.

18  Zhang D, Wang D. Relation Classification via Recurrent Neural Network. *arXiv:150801006 [cs]* Published Online First: 5 August 2015.http://arxiv.org/abs/1508.01006 (accessed 24 Jan 2018).

19  Chorowski JK, Bahdanau D, Serdyuk D, *et al.* Attention-based models for speech recognition. In: *Advances in neural information processing systems*. 2015. 577–585.

20  Zhou P, Shi W, Tian J, *et al.* Attention-based bidirectional long short-term memory networks for relation classification. In: *Proceedings of the 54th Annual Meeting of the Association for Computational Linguistics (Volume 2: Short Papers)*. 2016. 207–212.

21  Luo Y, Cheng Y, Uzuner Ö, *et al.* Segment convolutional neural networks (Seg-CNNs) for classifying relations in clinical notes. *Journal of the American Medical Informatics Association* 2018;**25**:93–98.

22  Li Y, Jin R, Luo Y. Classifying relations in clinical narratives using segment graph convolutional and recurrent neural networks (Seg-GCRNs). *Journal of the American Medical Informatics Association* 2019;**26**:262–268.

23  Li H, Tang B, Chen Q, *et al.* HITSZ_CDR: an end-to-end chemical and disease relation extraction system for BioCreative V. *Database* 2016;**2016**.

24  Song L, Zhang Y, Gildea D, *et al.* Leveraging dependency forest for neural medical relation extraction. *arXiv preprint arXiv:191104123* 2019.

25  Li F, Du J, He Y, *et al.* Time event ontology (TEO): to support semantic representation and reasoning of complex temporal relations of clinical events. *Journal of the American Medical Informatics Association* 2020.

26  Liu S, Wang L, Chaudhary V, *et al.* Attention neural model for temporal relation extraction. In: *Proceedings of the 2nd Clinical Natural Language Processing Workshop*. 2019. 134–139.

27  Vaswani A, Shazeer N, Parmar N, *et al.* Attention is all you need. In: *Advances in neural information processing systems*. 2017. 5998–6008.

28  Kenton JDM-WC, Toutanova LK. BERT: Pre-training of Deep Bidirectional Transformers for Language Understanding.

29  Shi P, Lin J. Simple BERT models for relation extraction and semantic role labeling. *arXiv preprint arXiv:190405255* 2019.

30  Xue K, Zhou Y, Ma Z, *et al.* Fine-tuning BERT for Joint Entity and Relation Extraction in Chinese Medical Text. *arXiv preprint arXiv:190807721* 2019.

31  Alt C, Hübner M, Hennig L. Improving relation extraction by pre-trained language representations. *arXiv preprint arXiv:190603088* 2019.

32  Uzuner Ö, South BR, Shen S, *et al.* 2010 i2b2/VA challenge on concepts, assertions, and relations in clinical text. *Journal of the American Medical Informatics Association* 2011;**18**:552–556.

33  Stubbs A, Filannino M, Soysal E, *et al.* Cohort selection for clinical trials: n2c2 2018 shared task track 1. *Journal of the American Medical Informatics Association* 2019;**26**:1163–1171.

34  Macarron R, Banks MN, Bojanic D, *et al.* Impact of high-throughput screening in biomedical research. *Nature reviews Drug discovery* 2011;**10**:188–195.

35  Greeley J, Jaramillo TF, Bonde J, *et al.* Computational high-throughput screening of electrocatalytic materials for hydrogen evolution. *Nature materials* 2006;**5**:909–913.

36  Hripcsak G, Albers DJ. Next-generation phenotyping of electronic health records. *Journal of the American Medical Informatics Association* 2013;**20**:117–121.

37  Yu S, Ma Y, Gronsbell J, *et al.* Enabling phenotypic big data with PheNorm. *Journal of the American Medical Informatics Association* 2017;**25**:54–60.

38  Liao KP, Sun J, Cai TA, *et al.* High-throughput multimodal automated phenotyping (MAP) with application to PheWAS. *Journal of the American Medical Informatics Association* 2019;**26**:1255–1262.

39  Mintz M, Bills S, Snow R, *et al.* Distant supervision for relation extraction without labeled data. In: *Proceedings of the Joint Conference of the 47th Annual Meeting of the ACL and the 4th International Joint Conference on Natural Language Processing of the AFNLP: Volume 2-Volume 2*. Association for Computational Linguistics 2009. 1003–1011.

40  Zeng D, Liu K, Chen Y, *et al.* Distant supervision for relation extraction via piecewise convolutional neural networks. In: *Proceedings of the 2015 Conference on Empirical Methods in Natural Language Processing*. 2015. 1753–1762.

41  Ji G, Liu K, He S, *et al.* Distant supervision for relation extraction with sentence-level attention and entity descriptions. In: *Thirty-First AAAI Conference on Artificial Intelligence*. 2017.

42  Lin Y, Li Y, Lu K, *et al.* Long-distance disorder-disorder relation extraction with bootstrapped noisy data. *Journal of Biomedical Informatics* 2020;:103529.

43  Zhao D, Weng C. Combining PubMed knowledge and EHR data to develop a weighted bayesian network for pancreatic cancer prediction. *Journal of biomedical informatics* 2011;**44**:859–868.

44  Liu Y, Ge T, Mathews KS, *et al.* Exploiting task-oriented resources to learn word embeddings for clinical abbreviation expansion. *arXiv preprint arXiv:180404225* 2018.

45  Munkhdalai T, Liu F, Yu H. Clinical relation extraction toward drug safety surveillance using electronic health record narratives: classical learning versus deep learning. *JMIR public health and surveillance* 2018;**4**:e29.

46  Phansalkar S, South BR, Hoffman JM, *et al.* Looking for a needle in the haystack? A case for detecting adverse drug events (ADE) in clinical notes. In: *AMIA... Annual Symposium proceedings. AMIA Symposium*. 2007. 1077–1077.

47  Iqbal E, Mallah R, Jackson RG, *et al.* Identification of adverse drug events from free text electronic patient records and information in a large mental health case register. *PloS one* 2015;**10**.

48  Aramaki E, Miura Y, Tonoike M, *et al.* Extraction of adverse drug effects from clinical records. *MedInfo* 2010;**160**:739–743.

49  Wang G, Jung K, Winnenburg R, *et al.* A method for systematic discovery of adverse drug events from clinical notes. *Journal of the American Medical Informatics Association* 2015;**22**:1196–1204.

50  Banda JM, Evans L, Vanguri RS, *et al.* A curated and standardized adverse drug event resource to accelerate drug safety research. *Scientific data* 2016;**3**:160026.

51  Gligorijevic D, Stojanovic J, Djuric N, *et al.* Large-scale discovery of disease-disease and disease-gene associations. *Scientific reports* 2016;**6**:1–12.

52  Gottesman O, Kuivaniemi H, Tromp G, *et al.* The electronic medical records and genomics (eMERGE) network: past, present, and future. *Genetics in Medicine* 2013;**15**:761–771.

53  McCarty CA, Chisholm RL, Chute CG, *et al.* The eMERGE Network: a consortium of biorepositories linked to electronic medical records data for conducting genomic studies. *BMC medical genomics* 2011;**4**:13.

54  Lee H-J, Xu H, Wang J, *et al.* UTHealth at SemEval-2016 task 12: an end-to-end system for temporal information extraction from clinical notes. In: *Proceedings of the 10th International Workshop on Semantic Evaluation (SemEval-2016)*. 2016. 1292–1297.

55  Leeuwenberg A, Moens MF. Structured learning for temporal relation extraction from clinical records. In: *Proceedings of the 15th Conference of the European Chapter of the Association for Computational Linguistics: Volume 1, Long Papers*. 2017. 1150–1158.

56  Schuyler PL, Hole WT, Tuttle MS, *et al.* The UMLS Metathesaurus: representing different views of biomedical concepts. *Bulletin of the Medical Library Association* 1993;**81**:217.

57  Bizer C, Lehmann J, Kobilarov G, *et al.* DBpedia-A crystallization point for the Web of Data. *Web Semantics: science, services and agents on the world wide web* 2009;**7**:154–165.

58  Vrandečić D, Krötzsch M. Wikidata: a free collaborative knowledgebase. *Communications of the ACM* 2014;**57**:78–85.

59  Lee J, Yoon W, Kim S, *et al.* BioBERT: a pre-trained biomedical language representation model for biomedical text mining. *Bioinformatics* 2020;**36**:1234–1240.

60  Beam AL, Kompa B, Fried I, *et al.* Clinical concept embeddings learned from massive sources of multimodal medical data. *arXiv preprint arXiv:180401486* 2018.

61  Laptop Diseases Database Ver 2.0; Medical lists and links Diseases Database. http://www.diseasesdatabase.com/ (accessed 16 Aug 2020).

62  Latest Medical News, Clinical Trials, Guidelines - Today on Medscape. https://www.medscape.com/ (accessed 24 May 2020).

63  Malani PN. Harrison's principles of internal medicine. *Jama* 2012;**308**:1813–1814.

64  Humes HD, DuPont HL, Gardner LB. *Kelley's textbook of internal medicine*. Lippincott Williams & Wilkins 2000.

65  Sabiston DC, Townsend CM, Beauchamp RD. *Sabiston textbook of surgery: the biological basis of modern surgical practice*. WB Saunders 2001.

66  Clark ML, Kumar P. *Kumar and Clark's clinical medicine*. 2017.

67  Zeng D, Liu K, Lai S, *et al.* Relation Classification via Convolutional Deep Neural Network. In: *COLING*. 2014. 2335–2344.

68  Brown TB, Mann B, Ryder N, *et al.* Language models are few-shot learners. *arXiv preprint arXiv:200514165* 2020.

# Supplementary Material

**Parameter settings**

Table 1: Hyper-parameters of the proposed model.

| parameter | Value |
|---|---|
| $d_e$ | 128 |
| $d_r$ | 200 |
| $d_k$ | 1000 |
| $d_c$ | 100 |
| $n_s$ | 10 |
| learning rate | $4 \times 10^{-4}$ |
| $l2$ penalty | $1 \times 10^{-7}$ |

**Sentence statistics**

Table 2: The number of extracted positive sentences for each relation from each source

| | Wikipedia | Textbooks | Medscape | UpToDate | PubMed |
|---|---|---|---|---|---|
| DDx | 6483 | 3120 | 14156 | 63108 | 269212 |
| MC | 7420 | 1984 | 9853 | 43739 | 153403 |
| MBCB | 4013 | 1576 | 5360 | 38191 | 136719 |
| IN | 3524 | 511 | 4408 | 15512 | 166292 |

**DISO-DISO results for each relation**

Table 3. Detailed DISO-DISO results of Models on Dataset 1

| Model | Acc. on All Samples | Acc. on Positive Samples | DDx Recall | DDx Precision | DDx F1 Score | MC Recall | MC Precision | MC F1 Score | MBCB Recall | MBCB Precision | MBCB F1 Score |
|---|---|---|---|---|---|---|---|---|---|---|---|
| Naïve Bayes | 0.509 | 0.075 | 0.132 | 0.697 | 0.222 | 0.042 | 0.597 | 0.079 | 0.004 | 0.333 | 0.008 |
| SVM | 0.406 | 0.330 | 0.420 | 0.332 | 0.371 | 0.288 | 0.268 | 0.278 | 0.209 | 0.210 | 0.209 |
| CNN | 0.755 | 0.549 | 0.657 | 0.783 | 0.714 | 0.487 | 0.737 | 0.587 | 0.415 | 0.655 | 0.508 |
| Bi-GRU | 0.828 | 0.678 | 0.724 | 0.794 | 0.757 | 0.743 | 0.662 | 0.700 | 0.523 | 0.785 | 0.628 |
| +attention+article structures | 0.826 | 0.725 | 0.817 | 0.702 | 0.755 | 0.707 | 0.681 | 0.694 | 0.574 | 0.697 | 0.629 |
| +CUI info | 0.896 | 0.830 | 0.844 | 0.842 | 0.843 | 0.840 | 0.803 | 0.821 | 0.794 | 0.796 | 0.795 |
| CUI only | 0.785 | 0.691 | 0.690 | 0.715 | 0.702 | 0.688 | 0.750 | 0.718 | 0.696 | 0.753 | 0.723 |
| BERT | 0.895 | 0.803 | 0.815 | 0.817 | 0.816 | 0.736 | 0.872 | 0.798 | 0.856 | 0.737 | 0.792 |
| BERT+CUI info | 0.934 | 0.876 | 0.907 | 0.872 | 0.889 | 0.853 | 0.895 | 0.874 | 0.842 | 0.885 | 0.863 |
| BioBERT+CUI info | 0.947 | 0.899 | 0.932 | 0.885 | 0.908 | 0.888 | 0.914 | 0.901 | 0.852 | 0.918 | 0.883 |

Table 4. Detailed DISO-DISO results of Models on Dataset 2

| Model | Acc. on All Samples | Acc. on Positive Samples | DDx Recall | DDx Precision | DDx F1 Score | MC Recall | MC Precision | MC F1 Score | MBCB Recall | MBCB Precision | MBCB F1 Score |
|---|---|---|---|---|---|---|---|---|---|---|---|
| Naïve Bayes | 0.593 | 0.289 | 0.417 | 0.778 | 0.543 | 0.273 | 0.534 | 0.361 | 0.059 | 0.471 | 0.105 |
| SVM | 0.566 | 0.322 | 0.468 | 0.567 | 0.513 | 0.224 | 0.378 | 0.278 | 0.151 | 0.270 | 0.209 |
| CNN | 0.761 | 0.585 | 0.712 | 0.774 | 0.742 | 0.528 | 0.701 | 0.602 | 0.652 | 0.404 | 0.499 |
| Bi-GRU | 0.821 | 0.690 | 0.668 | 0.839 | 0.744 | 0.761 | 0.667 | 0.711 | 0.650 | 0.710 | 0.679 |
| +attention+article | 0.839 | 0.731 | 0.748 | 0.795 | 0.771 | 0.754 | 0.679 | 0.714 | 0.671 | 0.663 | 0.667 |

| | | | | | | | | | | |
|---|---|---|---|---|---|---|---|---|---|---|
| structures | | | | | | | | | | |
| +CUI info | 0.900 | 0.833 | 0.836 | 0.874 | 0.854 | 0.861 | 0.779 | 0.818 | 0.796 | 0.819 | 0.807 |
| CUI only | 0.782 | 0.724 | 0.735 | 0.716 | 0.725 | 0.712 | 0.751 | 0.731 | 0.716 | 0.747 | 0.731 |
| BERT | 0.905 | 0.849 | 0.829 | 0.886 | 0.857 | 0.909 | 0.794 | 0.848 | 0.818 | 0.840 | 0.829 |
| BERT+CUI info | 0.933 | 0.888 | 0.893 | 0.901 | 0.897 | 0.907 | 0.874 | 0.890 | 0.857 | 0.901 | 0.879 |
| BioBERT+CUI info | 0.941 | 0.911 | 0.929 | 0.911 | 0.920 | 0.893 | 0.907 | 0.900 | 0.900 | 0.897 | 0.898 |

Table 5. Detailed DISO-DISO results of Models on Dataset mix

| Model | Acc. on All Samples | Acc. on Positive Samples | DDx Recall | DDx Precision | DDx F1 Score | MC Recall | MC Precision | MC F1 Score | MBCB Recall | MBCB Precision | MBCB F1 Score |
|---|---|---|---|---|---|---|---|---|---|---|---|
| Naïve Bayes | 0.504 | 0.154 | 0.252 | 0.736 | 0.375 | 0.109 | 0.575 | 0.184 | 0.017 | 0.286 | 0.031 |
| SVM | 0.483 | 0.317 | 0.427 | 0.437 | 0.431 | 0.223 | 0.328 | 0.266 | 0.215 | 0.263 | 0.236 |
| CNN | 0.737 | 0.557 | 0.685 | 0.768 | 0.724 | 0.526 | 0.662 | 0.586 | 0.346 | 0.669 | 0.456 |
| Bi-GRU | 0.822 | 0.698 | 0.780 | 0.760 | 0.770 | 0.667 | 0.716 | 0.691 | 0.577 | 0.771 | 0.660 |
| +attention+article structures | 0.819 | 0.728 | 0.759 | 0.763 | 0.761 | 0.777 | 0.639 | 0.701 | 0.610 | 0.660 | 0.634 |
| +CUI info | 0.894 | 0.835 | 0.865 | 0.838 | 0.851 | 0.822 | 0.834 | 0.828 | 0.793 | 0.795 | 0.794 |
| CUI only | 0.780 | 0.703 | 0.718 | 0.723 | 0.720 | 0.702 | 0.768 | 0.734 | 0.677 | 0.745 | 0.709 |
| BERT | 0.904 | 0.840 | 0.830 | 0.873 | 0.851 | 0.895 | 0.813 | 0.852 | 0.799 | 0.852 | 0.825 |
| BERT+CUI info | 0.928 | 0.881 | 0.872 | 0.909 | 0.890 | 0.893 | 0.880 | 0.887 | 0.882 | 0.844 | 0.863 |
| BioBERT+CUI info | 0.944 | 0.909 | 0.918 | 0.918 | 0.918 | 0.932 | 0.903 | 0.917 | 0.864 | 0.899 | 0.881 |